\pdfoutput=1
\documentclass[11pt]{article}

\usepackage{acl}

\usepackage{times}
\usepackage{latexsym}

\usepackage[T1]{fontenc}
\usepackage[utf8]{inputenc}

\usepackage{microtype}

\usepackage{inconsolata}

\usepackage{graphicx}

\title{Looking Into The Past: The Impact of Context on Detecting Deceptive Statements}
\author{
  \textbf{Alex Clay\textsuperscript{1}},
  \textbf{Ernesto Jimenez-Ruiz \textsuperscript{1}},
  \textbf{Pranava Madhyastha\textsuperscript{1,2}} \\
  \textsuperscript{1}City St George's, University of London,
  \textsuperscript{2}Alan Turing Institute
 }
\begin{document}
\maketitle
\begin{abstract}
%idea is essentially detecting deception in a gamified social setting, w generalisation as detecting falsehoods in social situations

%og ver:
%Determining how the structuring of contextual information can impact the quality of deception detection, aiming to see or prove the importance of having he context and potential benefits of structuring it and maybe reducing it to the necessary bits (which saves on tokens for the llm)

Deception detection typically considers messages in isolation. Previous works in claim verification and role detection have displayed that additional information such as gameplay history is crucial to successful predictions. In this work, we demonstrate the value of preceding gameplay dialogue when applied to deceptive message detection within the hidden role game of Avalon. The findings show that LLMs can see improvement across a number of metrics, including accuracy by up to 31\% with the addition of gameplay context for the target message.

%while traditional models like roberta do not benefit, LLMs are able to leverage the information to greatly improve their performance, though the degree to which varies depending on how they are prompted. 

\end{abstract}

\section{Introduction}

Textual deception detection has often been reliant on linguistic cues~\cite{Hauch2015-ib} and detection without contextualizing the target~\cite{velutharambath-klinger-2023-unidecor}. However, in settings like hidden role games, real-world social interactions are represented in a controlled environment where previous interaction history is obtainable. Many investigations of deception in hidden role games focus on detecting player roles from the game's history \cite{ibraheem-etal-2022-putting, stepputtis-etal-2023-long, chuchro2022trainingassassinairesistance}, but have yet to apply this same technique on a message by message basis. Moreover, the game Avalon in particular provides the unique opportunity for evaluation, as it is not only the 'evil'  but also the 'good' players who have incentive to lie. 

We explore, in particular, the application of preceding message context to deceptive message detection to determine if similar techniques to that of role detection can have a more widely applicable use. Both traditional deception detection techniques and LLMs were evaluated with and without history. We determined that the effect of such context is highly dependent on both the model type and presentation of information, but has widely positive effects for LLMs with improvements reaching up to 31\% for accuracy.% 49\% for specificity and
\section{Background and Related Work}

\subsection{Avalon \& Hidden Role Games}

Games provide a unique lens through which social dynamics can be studied in a controlled setting~\cite{Szczepanska04072022}. Avalon: The Resistance is a social cooperative hidden role game where players must collaborate to select parties for quests, ideally without including evil players. At its core, the game focuses on determining who is on your team and who is trying to deceive you. \citeauthor{light2023from} provides a more comprehensive overview of the game.

A common task within Avalon and the similar hidden role games is role detection, which in many cases has included game dialog but not conducted investigation into the importance of its presence. \citeauthor{ibraheem-etal-2022-putting} utilized a fine tuned BERT model with the addition of previous utterances to detect roles within the similar game Mafia. \citeauthor{stepputtis-etal-2023-long} similarly investigates the addition of chat history for role detection with LLMs in Avalon, while \citeauthor{chuchro2022trainingassassinairesistance} used an SVM on Avalon games, where 3 missions succeeded, to determine the role of Merlin, who must remain anonymous or risk losing the game for the 'good' side. Our work differs in that we aim to detect deception in a message rather than a role, which accounts for the unique factor of that both types of players might lie, increasing the difficulty of the task. We hypothesize that contextual information, especially that of dialogue history, could improve deception detection beyond even the scope of hidden role games.

%Additionally, the  we investigate that the technique of applying game history, typically used for role detection, could be more widely applicable to deception in social settings at large.

%+ mention other works where (we consider history

%While many approaches investigate detecting particular roles as a means for determining deception, in the game of Avalon, it is not just the 'evil' players who are incentivized to lie, highlighting the necessity of detecting deception on a message by message scale. 

%More recently, many aspects to evaluating hidden role gameplay agents have been turned over to LLMs [*]. However, this leads to the question of whether such approaches are as effective as more traditional methods, and whether providing the game's historical context of a message improves the ability to determine whether a message is deceptive. 

\subsection{Deception Detection}

Deception is tied closely to intention and belief \cite{10.5555/3666122.3666230} and can be abstractly considered to be instances where an individual is intentionally led to believe something that the deceiver does not believe to be true. 

Detection of deception in NLP has often relied on linguistic cues \cite{Zhou2004, Hauch2015-ib} and feature analysis \cite{mihalcea-strapparava-2009-lie, velutharambath-klinger-2023-unidecor, deruiter2019mafiascumdatasetlargetext}. In many cases, instances have been considered individually despite being derived from settings where additional information would provide necessary context. The Unidecor \cite{velutharambath-klinger-2023-unidecor} aggregated dataset, for instance, compiles a number of textual deception detection datasets, but does not retain order for those such as the Mafiascum dataset \cite{deruiter2019mafiascumdatasetlargetext} within which temporal factors could be relevant. 
More recently, LLMs have become prevalent within deception detection. \citeauthor{hazra-majumder-2024-tell} utilized LLMs to detect deception by comparing statements against an objective truth, to predict which contestant was truthful, and was able to identify accurate language cues. Similar tasks like fake news detection have also effectively employed LLMs. \citeauthor{Loconte2023} found that larger sizes of the FLAN-T5 model were more effective at detecting verbal lies. This was supported by \citeauthor{sallami2024deceptiondetectiondualroles} which also found larger models to be more effective in fake news detection, but highlighted that the training methods were equally important.

Providing additional context to a LLM is a common practice  in many tasks such as RAG \cite{10.5555/3648699.3648950} and claim verification \cite{momii-etal-2024-rag}. More specifically, \cite{10.1145/3643651.3659898} uses RAG for domain agnostic deception detection, highlighting the possibility for generalization in a task previously considered to be domain dependent \cite{velutharambath-klinger-2023-unidecor}.
 %Additionally, the supplementary information is not always text, as  \citeauthor{info16040297} combined LLMs with graph based representations to for detecting fake news, in order to capture more relationships within the texts. 
Within hidden role games, \citeauthor{ibraheem-etal-2022-putting} used language within the game of Mafia to identify deceptive players, while  \citeauthor{Yoo2024} applied LLMs. \citeauthor{Yoo2024} found that these methods were more effective than BERT like ones. The data was provided until a certain in-game day and the LLM was tasked with predicting which players were the mafia given partial record of the game. However, in the game Mafia, players are eliminated with each day cycle, which changes the likelihood of choosing a member of the mafia as the game continues. This is avoided in Avalon, as all players remain throughout the game. %In the game of Avalon, deception detection is more fine grained as the good side possesses a player who is actively incentivized to lie about their role.
We apply this same technique of integrating preceding gameplay dialogue information as a novel means for improving deception detection within social settings, like cooperative games, where there is an order and structure to interactions. 

\section{Methods}
Though the use of contextual information, such as game history, is crucial in tasks like role detection, with similar applications in fake news detection and claim verification, but has not been widely investigated within the frame of message specific deception detection. 

%While derived from methods for role detection, this approach is unique in its incorporation of historical context. Additionally, the use of these structures has been applied to claim verification and role detection, but there is novelty in its application to gamified deception detection. 
In order to discern the effect of such information, we considered more traditionally used models, as well as newer state-of-the-art models. The traditional deception detection model formats were used as a point of comparison, as well as to assess their ability to perform with the addition of preceding gameplay dialogue, which is generally not associated with such formats. The ordering structure for the game information was derived from \cite{ibraheem-etal-2022-putting}, where BERT was finetuned with prior game conversation preceding the utterance for prediction. 

To accommodate the necessity of prompting for Gemini, we tested prompt variations,  adapted from works in RAG \cite{10.5555/3648699.3648950} and claim verification for the FEVER task \cite{momii-etal-2024-rag} where additional information is used to predict one of three labels for a given claim. We conduct a number of ablations to understand how prompt variations and temperature impact generation, and to ascertain whether the traditional models benefit from prompt-formatted information.  

\section{Experimental set up}
\subsection{Avalon Dataset}
The test dataset is comprised of 20 games of Avalon from \citealp{stepputtis-etal-2023-long}, totaling 2384 messages. We retained the order of messages and player designation for each game, and converted the deception type category into a binary true/deceptive label. The versions including history were supplied with all messages in the game preceding the target message, for example: \texttt{Game History: "message A, message B" Message to predict: "message c"}. With each message, the player designation was included. 

%and was reformatted, retaining the order of messages and a binary label indicating deception, added by humans in the original work. Though the original dataset provided types of deception, this was not retained for this task. For the versions incorporating history, within a given game all messages prior to the target message were combined to create the game history to that point. Player ids were retained as part of each message, and to focus soley on communication voting was not includedr.
%\begin{figure}

%    \centering
%    \includegraphics[width=0.75\textwidth]{fig.jpeg}
 %   \caption{**}
 %   \label{fig}
%\end{figure}
%To compare the efficacy of these different approaches, using the human avalon dataset from  . The dataset provides messages within the game, noting deception where present. Though the dataset provides types of deception, this has been used in a binary format of whether there is deception or not. 

%Crucially, the deception labels on each utterance allow for the detection of deception rather than role in a hidden role game. This is of particular importance, as not every utterance by an 'evil' player will be deceptive, and not all utterances of a 'good' player will be true. In the game of Avalon, this is especially relevant as the 'good' player Merlin has incentive to be deceptive in order to protect the secrecy of their role. 

\subsection{Models Considered}
In order to represent more traditional methods for detecting deception, we selected Roberta, ModernBert, and an SVM. Roberta \cite{liu2019robertarobustlyoptimizedbert} has precedent in \citeauthor{velutharambath-klinger-2023-unidecor}, with other variants of BERT also commonly used for comparison \cite{hazra-majumder-2024-tell, 10.1145/3643651.3659898}. ModernBert \cite{warner2024smarterbetterfasterlonger} is a newer version of the eponymous BERT model, and was selected due to its better performance and capability to handle more tokens. The exact models we used were roberta-base and ModernBERT-base available on Hugging Face.

Feature-based approaches are prevalent in many linguistically based deception detection techniques, which we represented with an SVM \cite{chuchro2022trainingassassinairesistance, deruiter2019mafiascumdatasetlargetext}. The features selected for the SVM were adapted from the Mafiascum code with modification to apply to the new datasets. The model used was a standard SVM and scaler from the scikit-learn library \cite{scikit-learn}, the features for which are listed in full in Appendix A
%It has even been noted, even after the advent of transformers that more sophisticated approaches aren't necessarily more effective.

%\footnote{https://huggingface.co/FacebookAI/roberta-base}
%\footnote{https://huggingface.co/answerdotai/ModernBERT-base}
The three models were finetuned, and trained in the case of the SVM, on the mafia game dataset from \citeauthor{ibraheem-etal-2022-putting}. As the data lacks a deceptive label,  we marked any utterance by a player with the role 'mafioso' with the 'deceptive' label. This method mimics how the Mafiascum dataset was integrated into the Unidecor combined deception dataset, and was adopted due to a lack of game datasets with individually labeled messages. The game of Mafia and Avalon are very similar in format, allowing for minimal discrepancies.

%Additionally, the games of Mafia and Avalon are established* to be very similar, which is anticipated to avoid the lack of transferability across deception domains*. 

%As the dataset is originally presented in a network layout, the data was re-arranged to be in order of the id, and organized by game, and each message was assigned a 'T' or 'D' corresponding with whether it was stated by a bystander ('T' for True) or mafioso ('D' for Deceptive). 

%Owing to the small size of the https://github.com/omonida/mafia-dataset/tree/main mafia dataset both models were trained using a 95/5 for training and evaluation. 
We selected Gemini 2.0 Flash for the LLM due to the high quality of its generation and low cost. LLMs have precedent for accepting large quantities of additional information to supplement inference, making it an ideal setting for evaluating the impact of historical information. Any instances where Gemini failed to return an answer with a decision containing True or Deceptive were marked as incorrect. 

%In order to fully evaluate the efficacy of adding historical information to aid in deception detection, it was necessary to employ the use of LLMs which are known for their incorporation of context information.  was used, owing to it's quality? and free quota limits?

\subsection{Metrics}
While accuracy is commonly used for this variety of task \cite{10.1145/3643651.3659898, velutharambath-klinger-2023-unidecor, info16040297}, we found that it does not account for unbalanced datasets and can be misleading. In fact, many datasets for textual deception detection are imbalanced \cite{velutharambath-klinger-2023-unidecor}, like our test set which possesses an 82/18 split. In this work we included accuracy (both balanced and unbalanced for comparability)and macro precision. We selected macro precision to account for the unbalanced dataset and provide a view of the models overall efficacy. Our code is provided in Appendix~D.

%The metrics most commonly used in deception detection tasks are Accuracy[] and F1[https://dl.acm.org/doi/pdf/10.1145/3643651.3659898 https://www.mdpi.com/2078-2489/16/4/297], occassionally accompanied by Recall and Precision [https://www.mdpi.com/2078-2489/16/4/297]. However, accuracy can be misleading especially on unbalanced datasets, which comprise much of those avaliable for deception detection[*]. Matthews CC, conversely, accounts for unbalanced datasets and gives a clearer view of a models efficacy overall. 

\section{Experiments and Results}

Our findings in Table \ref{tab:resMetrics} show that the introduction of history generally improves Gemini's performance. Though a only small increase in macro precision and balanced accuracy, accuracy showed significant improvement, indicating that Gemini preformed well on the true label predictions. However, we notice that as the dataset was imbalanced in favor of the true label, it is possible that having gameplay dialogue data predominantly of one type skewed the latent patterns Gemini may have detected in one direction. 

However, as we anticipated, the more traditional approaches did not improve with the added information. While Roberta remained consistent in only predicting the true label, as ModernBert and the SVM did without history, the additional text actually reduced performance. Though the metrics reflect poorly, it is an interesting point that the history caused ModernBert and the SVM to begin predicting a small portion of deceptive labels (though not very well), perhaps indicating that the dialogue history may have simply acted as noise to these models. 

\begin{table}[h]
  \centering

  %\resizebox{.46\textwidth}{!}{
  {\footnotesize
  \begin{tabular}{lllllll}
    \hline
    \textbf{Model} & \textbf{Accuracy}   & \textbf{\begin{tabular}[c]{@{}c@{}}Balanced\\Accuracy\end{tabular}} & \textbf{\begin{tabular}[c]{@{}c@{}}Macro\\Precision\end{tabular}}\\\hline
     \ ModernBert          & 0.820 &  0.500 & 0.410\\
    \ ModernBert w/H    & 0.727 &  0.500&0.482\\
    \ Roberta          & 0.820 & 0.486&0.410\\
    \ Roberta  w/H        & 0.820 &   0.50&0.410\\
    \ SVM        & 0.820 & 0.50&0.410\\
     \ SVM w/H          & 0.815 & 0.467&0.410\\
    \ Gemini    & 0.378  &          0.520 & 0.514\\
    \ Gemini w/H   & 0.695 &   0.525& 0.521\\\hline
   % \ Gemini F  & 0.638 & 0.427 & 0.465  &         .533 &0.520&0.051 \\
   % \ Gemini F w/H   & 0.575 & 0.504 & 0.517 &          .540 &0.523 &0.061\\

%E is combo F is json
  %  \ SVM (Unidecor)     & {81.87}    &     Feature Based & \\
  %  \ Majority Class     & {82.04}    &     -  &  - & - \\\hline
  \end{tabular}
  
}
  \caption{Outcome of evaluation over \citeposs{stepputtis-etal-2023-long} human Avalon gameplay dataset. w/H denotes the model was provided with the game's dialogue data prior to the message.}
  \label{tab:resMetrics}
\end{table}

%evaluated the traditional approaches used in previous works and found that the models preformed poorly, generally only predicting the true label. The addition of history slightly reduced their performance by leading them to predict some deceptive labels(albeit usually incorrectly). We additionally fine tuned a separate version of each model %This had little impact though ModernBert preformed extremely poorly with around .17 accuracy compared with the .82 from always predicting true. A full table of these results is available in Appendix B. 

\begin{table}[h]
  \centering
  %\resizebox{.46\textwidth}{!}{
  {\footnotesize
  \begin{tabular}{lllllll}
    \hline
    \textbf{Model}  & \textbf{Accuracy}  &  \textbf{\begin{tabular}[c]{@{}c@{}}Balanced\\Accuracy\end{tabular}} &\textbf{\begin{tabular}[c]{@{}c@{}}Macro\\Precision\end{tabular}} \\\hline
     \ ModernBert & 0.180 &0.50& 0.090\\
     \ ModernBert w/H & 0.179 &0.498& 0.256\\
     \ Roberta & 0.820&0.50& 0.410\\
     \ Roberta  w/H & 0.820&0.50& 0.410\\
     \ SVM  &   0.815& 0.497&0.410  \\
     \ SVM w/H  & 0.820 & 0.50&0.410\\\hline

%E is combo F is json
  %  \ SVM (Unidecor)     & {81.87}    &     Feature Based & \\
  %  \ Majority Class     & {82.04}    &     -  &  - & - \\\hline
  \end{tabular}
}
  \caption{Outcome of fine tuning and evaluating traditional models with the same prompt structure as Gemini.}
  \label{tab:resMetricspromptfmt}
\end{table}

In order to better understand the full impact of adding dialogue history to the more traditional approaches, we also fine tuned and evaluated them using the exact same layout (prompt and all) as Gemini received, as well as  in the evaluation, the results for which are in Table \ref{tab:resMetricspromptfmt}. Though Roberta and the SVM saw essentially no change, ModernBert's results completely reversed, reaching a perfect score on the deceptive labels without history, and a near perfect with. This indicates that ModernBert, with its much larger context window may be significantly impacted by the presence of a prompt or additional structure in the presence of information. However, the addition of history still degraded performance, possibly due to the mixed content of true and deceptive in the gameplay dialogue introducing noise. Additionally, this more structured version contained an example of a deceptive message, which may have skewed ModernBert's prediction as the prompt discussed deceptive messages more than true ones.

\paragraph{Ablations}
%Three primary ablations were conducted, variations in the prompt, the removal of the example, and variations in temperature.

In order to maximize the effectiveness of the LLM, we carried out a number of prompt variations on a subsample of the test dataset, the outcomes for which are available in full in Appendix C. We derived prompts from \citeauthor{10.5555/3648699.3648950} and \citeauthor{momii-etal-2024-rag} due to the similarity of the domains. We found that a more narrative structure improved prediction of deceptive labels to an equivalent degree that a formal data structure improved true labels, showing that the labels could potentially have an inverse relationship, which is mirrored by the results of ModernBert. The prompt used in Table \ref{tab:resMetrics} is a combination of the prior two. These findings show that Gemini is consistent with the findings that some GPT based models are significantly impacted by prompt format \cite{he2024doespromptformattingimpact}. 

%

%E and F were two combinations of C and D in the interest of capturing the benefits of both variations. However, it became apparent that there may be a trade-off between detecting true and deceptive labels, as increasing one consistently reduced the other. %and that a multi-model set up with optimizations for each label may be the optimal solution.

We also produced an example message that was not present in the dataset, but maintained the hallmarks of one of the deceptive messages,  through prompting the Gemini model to create a new message based on all of the deceptive messages in the test data set. Interestingly, the removal of the example improved metrics on the sample, however aided in the full set as it favored true labels.  %The resulting message was: \texttt{ "i'm good, so i'll vote yes on this party. i'm not sure about player-3, but i'm not sure about anyone at this point."}

Similar works investigating the supplement of information to an LLM investigated differing temperatures during generation \cite{Yoo2024}. As higher temperature is often considered to be less probabilistic in generation, we considered that this could have a positive impact on the prediction of deception. Prompt A was evaluated with history on the sample data across 0 to 1.0 temperatures in .2 increments. We found that the results held consistent with recent research, that between 0 and 1, changes in temperature are not statistically significant to problem solving situations \cite{renze-2024-effect}. A table with these results is available in Appendix~B.

\paragraph{The Task}
Avalon presents a unique difficulty as not only do all players stay the whole game, but lying is beneficial to both sides. The addition of the Merlin and Assassin characters causes deception to be critical to the success of both the 'good' and 'evil' team. Moreover, the task in of itself is difficult, as the authors ourselves struggled to accurately label the messages, highlighting this as a valuable challenge for both AI and humans alike. %\citeauthor{light2023from} highlights  
\section{Discussion}

While adding gameplay dialogue history is crucial to detecting player patterns in tasks of role deception, in per-message deception detection, it is also shown to improve prediction. It is possible that direct instruction to consider the behavior of the player uttering the message, perhaps an LLM would be more capable of utilizing the additional information particular to patterns of player actions. Additionally, it was unexpected that while Gemini generally improved overall predictions, became less adept at the prediction of the deception label. There could be a number of reasons for this, including potentially the structuring of the information as it was presented to the LLM. The reduction of performance in the BERT variants and SVM were largely anticipated due to their design. However, ModernBert's behavior with the introduction of the prompt and additional structure leaves questions about how the model behaves with prompting, something we did not specifically investigate.

%[++ based on what argument is made]
%This is consistent with the prompt variation study which showed that deceptive and true labeling could potentially have an inverse relationship. The changes between the more structured and more narrative prompt variations also indicate that information may potentially evoke different behavior within an LLM, possibly due to patterns learned from data at training time.

%Lastly, the BERT variant models and SVM held with expectation that their performance would not be optimized with additional information, 

%though it was anticipated that the addition of historical context might degrade performance of the BERT models due to their structure

\section{Conclusion}

 We explored the application of gameplay dialogue history information in detecting deception in messages within the social setting of a hidden role game. Our findings show that indeed the additional information improves an LLM's performance, though this is not true of SVMs or BERT variant models. These findings present an interesting direction in which textual deception detection could continue to improve in social settings, by leveraging the information preceding the target message. This shows that when it comes to hidden role games, perhaps the future of deception detection is truly in the past.

\bibliography{ref}

\appendix
\label{sec:appendix}
\section{SVM Features}
The features used are as follows: word count, or ratio, negative emotion ratio, anger ratio, sensory ratio, cognitive ratio, insight ratio, motion ratio, tentative ratio, not ratio (didn't, wasn't...), pronoun ration (first, second, and third respectively), quantifier ratio (all, some, many, etc...), sentence length, but ratio, token length, unique tokens, message count, message length, and the vector sum of the sentence.

\section{Temperature}
\begin{table}[h]
  \centering
  %\resizebox{.46\textwidth}{!}{
  {\footnotesize
  \begin{tabular}{llllllll}
    \hline
    \textbf{Temperature} & \textbf{Accuracy} & \textbf{\begin{tabular}[c]{@{}c@{}}Macro\\Precision\end{tabular}} & \textbf{\begin{tabular}[c]{@{}c@{}}Balanced\\Accuracy\end{tabular}}\\\hline
    \ 0.0     & 0.535 & 0.536 & 0.536\\
     \ 0.2    &0.535 &0.536 & 0.536\\
    \ 0.4     &0.523&0.523 &0.523\\
    \ 0.6     &0.525&0.525 & 0.526\\
    \ 0.8   &0.529&0.530 & 0.529\\
    \ 1.0     &0.510 & 0.511 &0.511\\\hline
  \end{tabular}
  }
  \caption{Evaluating the performance of varying temperatures, Prompt A with history was used as the configuration. }
  \label{tab:resTemp}
\end{table}

\section{Prompts}
%Prompt A (Original Prompt) promptOGfmt promptOGwHistoryfmt

\begin{table}[h]
  \centering
 % \resizebox{.46\textwidth}{!}{
 {\footnotesize
  \begin{tabular}{lllllll}
    \hline
    \textbf{Prompt} &\textbf{Accuracy}  &\textbf{\begin{tabular}[c]{@{}c@{}}Macro\\Precision\end{tabular}}& \textbf{\begin{tabular}[c]{@{}c@{}}Balanced\\Accuracy\end{tabular}}\\\hline
    %\ promptOGfmt    & {0.615}      & 0.712    &  1.0 & 0.683 & 0.538 & 0.553 \\
   % promptOGwHistoryfmt
     \   A    & 0.535 & 0.536 & .536\\
     %promptwExamplefmt
    \  B    & 0.506 & 0.506 & .507\\
    %feverpromptstructfmt
    \     C  & 0.548 &0.549 & .548 \\
    %jsonpromptmessfmt
    \  D    & 0.548 & 0.549 &.548\\
    %combopromptfmt
    \  E    & 0.510 & 0.515 & .512\\\hline
    %jsonremix
   % \ F & 0.129  & 0.579 & 0.550 &0.565 & 0.565 & .511\\\hline
    %\ combopromptremixfmt     & {0.560}    &  0.699 & 1.0  & 0.583 & 0.533 & 0.537\\
    %\ combopromptremixthesequelfmt     & {0.549}    & 0.708   &  1.0 & 0.550 & 0.548 & 0.548 \\
   % \ jsonremixfmt    & {0.556}      & 0.720     & 1.0 & 0.550 & 0.565 & 0.563\\
   % \ jsonremixinspacefmt    & {0.597}      & 0.697    & 1.0 & 0.667 & 0.525 & 0.535 \\\hline
  \end{tabular}
  }
  \caption{Evaluation of prompt variations. The sample totaled 480 messages, about 20\% of the test dataset, with 240 D and 240 T.}
  \label{tab:promptchecks}
\end{table}

\subsection{Prompt A: Original Prompt}
\texttt{ """You are a helpful assistant that uses the chat between six players, player-1 to player-6, who play Avalon: The Resistance (a cooperative-competitive game) to identify which messages are true or deceptive. Within the game, there can be multiple evil players, who have incentive to be deceptive and lie about their roles and those of other players. There is also a good character Merlin, who has knowledge of evil players’ identities, and may lie to protect their identity. Given the following message, do you think it is true or deceptive? Please do not explain your answer, only provide the answer and nothing else."""}

\subsection{Prompt B: Adding an Example}
%Prompt B promptwExamplefmt

\texttt{"""You are a helpful assistant that uses the chat between six players, player-1 to player-6, who play Avalon: The Resistance (a cooperative-competitive game) to identify which messages are true or deceptive. Within the game, there can be multiple evil players, who have incentive to be deceptive and lie about their roles and those of other players. There is also a good character Merlin, who has knowledge of evil players’ identities, and may lie to protect their identity. An example of a deceptive message is:\\
"i'm good, so i'll vote yes on this party. i'm not sure about player-3, but i'm not sure about anyone at this point."\\
Given the following message, do you think it is true or deceptive? Please do not explain your answer, only provide the answer and nothing else."""}

\subsection{Prompt C: Fever challenge paper}
%Prompt C feverpromptstructfmt

\texttt{ """You will be given a message from the hidden role game Avalon. Your task is to determine whether this message is True or Deceptive. Your prediction must be based on the game history and given example. \\ Example of a deceptive message : \\ "i'm good, so i'll vote yes on this party. i'm not sure about player-3, but i'm not sure about anyone at this point." \\ The game history is as follows: \\ """ + x + """\\ The message to predict is: \\ """ + row['withName'] + """\\ Output Format: "True" or "Deceptive" """ }

\subsection{Prompt D: Json-like}
%Prompt D jsonpromptmess

\texttt{ """\{"Task" : "You will be given a message from the hidden role game Avalon. Your task is to determine whether this message is True or Deceptive. Your prediction must be based on the game history and given example.",  "Example of a deceptive message" : "i'm good, so i'll vote yes on this party. i'm not sure about player-3, but i'm not sure about anyone at this point." , "Game History" : """ + x + """, "Message to predict" : """ + row['withName'] + """,  "Output Format" : "True" or "Deceptive" \}""" }

\subsection{Prompt E: Combination C and D}
%Prompt E combopromptfmt

\texttt{"""\{"Task" : "You will be given a message from the hidden role game Avalon. Your task is to determine whether this message is True or Deceptive. Your prediction must be based on the game history and given example.", \\  "Example of a deceptive message" : "i'm good, so i'll vote yes on this party. i'm not sure about player-3, but i'm not sure about anyone at this point." , \\ "Game History" : """ + x + """,\\ "Message to predict" : """ + row['withName'] + """, \\ "Output Format" : "True" or "Deceptive" \}""" }

\section{Code}
Repository is available here: \url{https://github.com/alexclay42/history-paper}

\end{document}